\def\year{2020}\relax
\documentclass[letterpaper]{article} 
\usepackage{aaai20}  
\usepackage{times}  
\usepackage{helvet} 
\usepackage{courier}  
\usepackage[hyphens]{url}  
\usepackage{graphicx} 
\usepackage{amssymb,bm,mathrsfs}
\usepackage{amsfonts, amsmath}
\usepackage{algorithm}
\usepackage{paralist,booktabs}
\usepackage{multirow}
\usepackage{float}
\usepackage{algorithm}
\usepackage[noend]{algpseudocode}
\usepackage{color,xcolor}

\usepackage{stfloats} 
\newcommand{\citet}[1]{\citeauthor{#1}~\shortcite{#1}}

\newcommand{\citep}{\cite}

\title{Unsupervised Paraphrasing by Simulated Annealing}

\author{Xianggen Liu,$^1$ \ Lili Mou,$^2$ \ Fandong Meng,$^3$ \ Hao Zhou,$^4$ \ Jie Zhou,$^3$ \ Sen Song$^1$\\ $^1${Tsinghua University}, $^2${University of Alberta} $^3${WeChat AI, Tencent Inc},$^4${ByteDance AI Lab}
\\
liuxg16@mails.tsinghua.edu.cn, doublepower.mou@gmail.com, fandongmeng@tencent.com, \\
zhouhao.nlp@bytedance.com, withtomzhou@tencent.com, songsen@tsinghua.edu.cn
}

\begin{document}

\maketitle
\begin{abstract}
  Unsupervised paraphrase generation is a promising and important research topic in natural language processing. We propose UPSA, a novel approach that accomplishes Unsupervised Paraphrasing by Simulated Annealing. We model paraphrase generation as an optimization problem and propose a sophisticated objective function, involving semantic similarity, expression diversity, and language fluency of paraphrases. Then, UPSA searches the sentence space towards this objective by performing a sequence of local editing. Our method is unsupervised and does not require parallel corpora for training, so it could be easily applied to different domains. We evaluate our approach on a variety of benchmark datasets, namely, Quora, Wikianswers, MSCOCO, and Twitter. Extensive results show that UPSA achieves the state-of-the-art performance compared with previous unsupervised methods in terms of both automatic and human evaluations. Further, our approach  outperforms most existing domain-adapted supervised models, showing the generalizability of UPSA.\footnote{Code and data are available at: {https://github.com/anonymity-person/UPSA} (anonymized  during double-blind review).}
\end{abstract}

\section{Introduction}
Paraphrasing aims to restate one sentence  as another with the same meaning, but different wordings. 
It constitutes a corner stone in many NLP tasks,  such as question answering~\cite{mckeown1983paraphrasing}, information retrieval~\cite{knight2000statistics}, and dialogue systems~\cite{shah2018building}. 
However, automatically generating accurate and
different-appearing paraphrases is a still challenging research problem, due to the complexity of natural language. 

Conventional approaches~\cite{prakash2016neural,gupta2018deep} model the paraphrase generation as a supervised encoding-decoding problem, inspired by machine translation systems. Usually, such models require massive parallel samples for training. In machine translation, for example, the WMT 2014 English-German dataset contains 4.5M sentence pairs~\cite{wmt14}.

However, the training corpora for paraphrasing are usually small. The widely-used Quora dataset\footnote{https://www.kaggle.com/c/quora-question-pairs} only contains 140K pairs of paraphrases; constructing such human-written paraphrase pairs is expensive and labor-intensive. Further, existing paraphrase datasets are domain-specific: the Quora dataset only contains question sentences, and thus, supervised paraphrase models do not generalize well to new domains~\cite{zichao2019}.
On the other hand, researchers synthesize pseudo-paraphrase pairs by clustering news events~\cite{Barzilay-rules}, crawling tweets of the same topic~\cite{lan2017a}, or translating bi-lingual datasets~\cite{wieting2017paranmt}, but these methods typically yield noisy training sets, leading to low paraphrasing performance~\cite{li2017paraphrase}. 

As a result, unsupervised methods would largely benefit the paraphrase generation task if no parallel data are needed. With the help of deep learning, researchers are able to generate paraphrases by sampling from a neural network-defined probabilistic distribution, either in a continuous latent space~\cite{bowman2015generating} or directly in the word space~\cite{miao2018cgmh}. However, the meaning preservation and expression diversity of those generated paraphrases are less ``controllable'' in such probabilistic sampling procedures.

\begin{figure}[!t]
\centering
\includegraphics[width=1\linewidth]{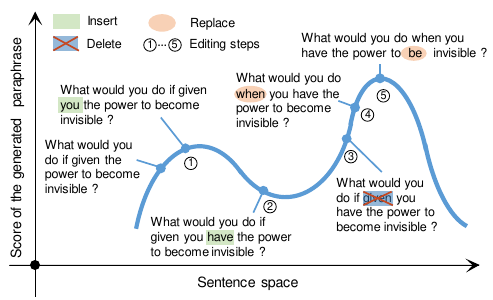}
	\caption{UPSA generates a paraphrase by a series of editing operations (i.e., insertion, replacement, and deletion). 
	At each step, UPSA proposes a candidate modification of the sentence, which is accepted or rejected according to a certain acceptance rate (only accepted modifications are shown).
Although sentences are discrete, we make an analogue in the continuous real $x$-axis where the distance of two sentences is roughly given by the number of edits.
}
\label{fig:upsa}
\end{figure}

To this end, we propose a novel approach to \textit{Unsupervised Paraphrasing by Simulated Annealing} (UPSA). 
Simulated annealing (SA) is a stochastic searching algorithm towards an objective function, which can be flexibly defined. In our work, we design a sophisticated objective function, considering semantic preservation, expression diversity, and language fluency of paraphrases.
SA searches towards this objective by performing a sequence of local editing steps, namely, word replacement, insertion, and deletion.
For each step, UPSA first proposes a potential editing, and then accepts or rejects the proposal based on sample quality. In general, a better sentence (higher scored in the objective) is always accepted, while a worse sentence is likely to be rejected, but could also be accepted (controlled by an annealing temperature). At the beginning, the temperature is usually high, and worse sentences are more likely to be accepted. This pushes the SA algorithm outside a local optimum. The temperature is cooled down as the optimization proceeds, making the model better settle down to some optimum.
Figure~\ref{fig:upsa} illustrates how UPSA searches an optimum in unsupervised paraphrase generation. 

We evaluate the effectiveness of our model on four paraphrasing datasets, namely, Quora, Wikianswers,  MSCOCO, and Twitter. Experimental results show that UPSA achieves a new state-of-the-art unsupervised performance in terms of both automatic metrics and human evaluation. Our unsupervised approach also outperforms most domain-adapted paraphrase generators. 

In summary, our contributions are as follows:
\begin{compactitem}

    \item  We propose the novel UPSA framework for Unsupervised Paraphrasing approach by Simulated Annealing.
    
   \item We design a searching objective function of SA that  not only considers language fluency and semantic similarity, but also explicitly models expression diversity between a paraphrase and the input.
    
    \item We propose a copy mechanism as one of our searching action during simulated annealing to address rare words. 
    
    \item We achieve the state-of-the-art performance on four benchmark datasets among all unsupervised paraphrase generators, largely reducing the performance gap between unsupervised and supervised paraphrasing. We outperform most domain-adapted paraphrase generators, and even a supervised one on the Wikianswers dataset.
    \end{compactitem}

\section{Related Work}
In early years, paraphrasing is typically accomplished by exploiting linguistic knowledge, including handcrafted rules~\cite{mckeown1983paraphrasing}, grammar structures~\cite{ellsworth2007mutaphrase,narayan2016paraphrase}, and shallow features~\cite{zhao2009application}. Recently, deep neural networks have become a prevailing approach for text generation~\cite{gupta2018deep}, where paraphrasing is often formulated as a supervised encoding-decoding problem, for example, stacked residual LSTM~\cite{prakash2016neural} and the Transformer model~\cite{wang2019a}.

\citet{zichao2019} learn paraphrasing at different levels of granularity (namely, sentence- and word-level paraphrasing), also in a supervised fashion. This achieves the state-of-the-art performance of paraphrase generation and is more generalizable to new domains.

Unsupervised paraphrasing is an emerging research direction in the field of NLP. The variational autoencoder (VAE) can be intuitively applied to  paraphrase generation in an unsupervised fashion, as we can sample sentences from a learned latent space~\cite{bowman2015generating}. But the generated sentences are less controllable and suffer from the error accumulation problem in VAE's decoding phase~\cite{miao2018cgmh}. 

\citet{roy-grangier-2019-unsupervised} introduce an unsupervised model based on vector-quantized autoencoders~\cite{van2017neural}. But their work mainly focuses on generating sentences for data argumentation instead of paraphrasing itself, and is not directly comparable.

\citet{miao2018cgmh} use Metropolis--Hastings sampling~(\citeyear{metropolis1953equation}) for constrained sentence generation, achieving the state-of-the-art unsupervised paraphrasing performance. The main difference between their work and ours is that we formulate paraphrasing as a stochastic searching problem. In addition, we define our searching objective involving not only semantic similarity and language fluency, but also the expression diversity; we further propose a copy mechanism in our searching process.

Recently, a few studies have applied editing-based approaches to sentence generation. \citet{guu2018generating} propose a heuristic delete-retrieve-generate approach as a component of a supervised sequence-to-sequence (Seq2Seq) model, but our UPSA is a mathematically inspired, unsupervised searching algorithm. \citet{dong2019editnts} learn the deletion and insertion operations for text simplification in a supervised way, where their groundtruth operations are obtained by some dynamic programming algorithm. Our editing operations (insertion, deletion, and replacement) are the searching actions of unsupervised simulated annealing.

Regarding discrete optimization/searching, a na\"ive approach is by hill climbing~\cite{edelkamp2011heuristic}, which is in fact a greedy algorithm. In NLP,  beam search (BS, \citeauthor{beam}~\citeyear{beam}) is widely applied to sentence generation. BS maintains a $k$-best list in a partially greedy fashion during left-to-right (or right-to-left) decoding~\cite{anderson2016guided}. By contrast, UPSA makes distributed modifications over the entire sentence. Moreover, UPSA is able to make use of the original sentence as an initial state of searching, whereas BS usually works in the decoder of a sequence-to-sequence model and is not applicable to unsupervised paraphrasing.

\section{Approach}

In this section, we present our novel UPSA framework that uses  simulated annealing (SA)  for unsupervised paraphrase generation. In particular, we first present the general SA algorithm, and then design our searching objective and searching actions (i.e., candidate sentence generator).

\subsection{The Simulated Annealing Algorithm}

Simulated Annealing (SA) is an effective and general metaheuristic of searching, especially for a large discrete or continuous space~\cite{Kirkpatrick671}. 

Let $\mathcal{X}$ be a (huge) search space of sentences, and $f(\mathrm x)$ be an objective function. The goal is to search for a sentence $\mathrm x$ that \textbf{maximizes} $f(\mathrm x)$. At a searching step $t$, SA keeps a current sentence $\mathrm x_t$, and proposes a new candidate $\mathrm x_*$ by local editing. If the new candidate is better scored by $f$, i.e., $f(\mathrm x_*)>f(\mathrm x_t)$, then SA accepts the proposal. Otherwise, SA tends to reject the proposal $x_*$, but may still accept it with a small probability $e^{\frac{f(\mathrm x_*)-f(\mathrm x_t)}{T}}$, controlled by an annealing temperature $T$. In other words, the probability of accepting the proposal is
\begin{align}
\label{eq:temp}
    p(\text{accept}|\mathrm x_*,\mathrm x_t, T) = \min\big(1, e^{\frac{f(\mathrm x_{*})-f(\mathrm x_{t})}{T}}\big).
\end{align}
If the proposal is accepted, $\mathrm x_{t+1}=\mathrm x_*$, or otherwise, $\mathrm x_{t+1}=\mathrm x_t$.

Inspired by the annealing in chemistry, the temperature $T$ is usually high at the beginning of searching, leading to a high acceptance probability even if $\mathrm x_*$ is worse than $\mathrm x_t$.  Then, the temperature is decreased gradually as the search proceeds. 
In our work, we adopt the linear annealing schedule, given by $T = \max(0, T_\text{init}- C\cdot t)$, where  $T_\text{init}$ is the initial temperature and $C$ is the decreasing rate. 

The high initial temperature of SA makes the algorithm less greedy compared with hill climbing, whereas the decreasing of temperature along the search process enables it to better settle down to a certain optimum. 

Theoretically, simulated annealing is guaranteed
to converge to the global optimum in a finite problem if the proposal and the temperature satisfy some mild conditions~\cite{granville1994simulated}.
{Although} such convergence may be slower than exhaustive search and the sentence space is, in fact, potentially infinite, simulated annealing is still a widely applied searching algorithm, especially for discrete optimization. Readers may refer to \citet{hwang1988simulated} for details of the SA algorithm. 

\subsection{Objective Function}
Simulated annealing maximizes an objective function, which can be flexibly specified in different applications.
In particular, our UPSA objective $f(\mathrm x)$ considers multiple aspects of a candidate paraphrase, including semantic preservation $f_\text{sem}$, expression diversity $f_\text{exp}$, and language fluency $f_\text{flu}$. Thus, our searching objective is to maximize \begin{align}
\label{eq:obj}
 f(\mathrm x) &= f_{\text{sem}}(\mathrm x,\mathrm x_0) \cdot f_{\text{exp}}(\mathrm x,\mathrm x_0)\cdot f_{\text{flu}}(\mathrm x),
\end{align}
where $\mathrm x_0$ is the input sentence. 

\textbf{Semantic Preservation.} A paraphrase is expected to capture all the key semantics of the original sentence. Thus, we leverage the cosine function of keyword embeddings to measure if the key focus of the candidate paraphrase is the same as the input.
We use \citet{rose2010automatic} to extract the keywords of the input sentence $\mathrm x_0$ and embed them by GloVE~\cite{pennington2014glove}. For each keyword, we find the closest word in the candidate paraphrase $\mathrm x_*$ in terms of cosine similarity. Our keyword-based semantic preservation score is given by the lowest cosine similarity among all keywords, i.e., the least matched keyword:
 \begin{align}
    f_{\text{sem,key}}(\mathrm x_{*},\mathrm x_0)=\min_{ e\in \text{keywords}(\mathrm x_0)}\max_j\{\cos(\bm w_{*,j},\bm e)\},
 \end{align}
where  $w_{*,j}$ is  the  $j$th word in the sentence $\mathrm x_*$; $e$ is an extracted keyword of $\mathrm x_0$. Bold letters indicate embedding vectors.

In addition to keyword embeddings, we also adopt a sentence-level similarity function, based on Sent2Vec embeddings~\cite{pagliardini2017unsupervised}.
Sent2Vec learns $n$-gram embeddings and computes the average of $n$-grams embeddings as the sentence vector. It has been shown significant improvements over other unsupervised sentence embedding methods in similarity evaluation tasks~\cite{pagliardini2017unsupervised}. 
Let ${\mathbf x}_*$ and  $\mathbf x_0$ be the Sent2Vec embeddings of the candidate paraphrase and the input sentence, respectively. Our sentence-based semantic preservation scoring function is $f_{\text{sim,sen}}(\mathrm x_{*},\mathrm x_0)=\cos({\mathbf x}_{*},{\mathbf x}_0)$.

To sum up, the overall semantic preservation scoring function of UPSA is given by
\begin{align}
    f_{\text{sem}}(\mathrm x_{*},\mathrm x_0)=  f_{\text{sem,key}}(\mathrm x_{*},\mathrm x_0)^P\cdot f_{\text{sem,sen}}(\mathrm x_{*},\mathrm x_0)^Q,
 \end{align}
where $P$ and $Q$ are hyperparameters, balancing the importance of the two factors. Here, we use power weights because the scoring functions are multiplicative.  

\textbf{Expression Diversity.}
The expression diversity scoring function computes the lexical difference of two sentences. We adopt a BLEU-induced function to penalize the repetition of the words and phrases in the input sentence:
\begin{align}
   f_{\text{exp}}(\mathrm x_{*},\mathrm x_0)=(1-\text{BLEU}(\mathrm x_*,\mathrm x_0))^S,
\end{align}
where the BLEU score~\cite{papineni2002bleu} computes a length-penalized geometric mean of $n$-gram precision ($n=1,\cdots,4$). $S$ coordinates the importance of $f_{\text{exp}}(\mathrm x_{t},\mathrm x_0)$ in the objective function (\ref{eq:obj}).

\textbf{Language Fluency.} Despite semantic preservation and expression diversity, the candidate paraphrase should be a fluent sentence by itself. We use a separately trained (forward) language model (denoted as $\overrightarrow{\text{LM}}$) to compute the likelihood of the candidate paraphrase as our fluency scoring function:
\begin{align}
   f_{\text{flu}}(\mathrm x_{*})=\prod_{k=1}^{k=l_*} p_{\overrightarrow{\text{LM}}}( w_{*,k}|  w_{*,1},\dots, w_{*,k-1}),
\end{align}
where $l_*$ is the length of $\mathrm x_*$ and $w_{*,1},\dots, w_{*,l}$ are words of $\mathrm x_*$.
Here, we use a dataset-specific language model, trained on  non-parallel sentences. Notice that a weighting hyperparameter is not needed for  $f_{\text{flu}}$, because the relative weights of different factors in Eqn.~(\ref{eq:obj}) are given by the powers in $f_\text{sem,key}$, $f_\text{sem,sen}$, and $f_\text{exp}$.

\subsection{Candidate Sentence Generator}
\label{sec:generation}

As mentioned, simulated annealing proposes a candidate sentence at each searching action, which is either accepted or rejected by Eqn.~(\ref{eq:temp}). Since each action yields a new sentence $\mathrm x_*$ from $\mathrm x_t$, we call it a \textit{candidate sentence generator}. While the proposal of candidate sentences does not affect convergence in theory (if some mild conditions are satisfied), it may largely influence the efficiency of SA searching.

In our work, we mostly adopt the word-level editing in \citet{miao2018cgmh} as our searching actions. At each step $t$, the candidate sentence generator randomly samples an editing position $k$ and an editing operation， namely, replacement, insertion, and deletion. For replacement and insertion, the candidate sentence generator also samples a candidate word.

Let the current sentence be $\mathrm x_t=(w_{t,1},\dots, w_{t,k-1},$ $w_k,w_{t,k+1}\dots, w_{t,l_t})$. If the replacement operation proposes a candidate word $w_*$ for the $k$th step, the resulting candidate sentence becomes $\mathrm x_*=(w_{t,1},\dots, w_{t,k-1},$ $ w_*,w_{t,k+1}\dots, w_{t,l_t})$. The insertion operation works similarly.

Here, the word is sampled from a probabilistic distribution, induced by the objective function (\ref{eq:obj}):
\small
 \begin{align}
     p(w_*|\cdot)&=
\frac{f_{\text{sim}}(\mathrm x_{*},\mathrm x_0) \cdot f_{\text{exp}}(\mathrm x_{*},\mathrm x_0) \cdot  f_{\text{flu}}(\mathrm x_*)}{Z},\\
     Z &= \sum_{w_{*}\in\mathcal{W}}f_{\text{sim}}(\mathrm x_{*},\mathrm x_0) \cdot f_{\text{exp}}(\mathrm x_{*},\mathrm x_0) \cdot  f_{\text{flu}}(\mathrm x_*) \label{eq:norm},
 \end{align}\normalsize
where $\mathcal{W}$ is the sampling vocabulary; $Z$ is known as the normalizing factor (noticing our scoring functions are nonnegative). We observe that sampling from such objective-induced distribution typically yields a meaningful candidate sentence, which enables SA to explore the search space more efficiently.

It is also noted that sampling a word from the entire vocabulary involves re-evaluating (\ref{eq:obj}) for each candidate word, and therefore, we also follow \citet{miao2018cgmh} and only sample from the top-$K$ words given by jointly considering a forward language model and backward language model. The replacement operator, for example, suggests the top-$K$ words vocabulary by 
\begin{align}
    \mathcal{W}_{t,\text{replace}}= \operatorname{top-}K_{w_*} \Big[p_{\overrightarrow{\text {LM}}}(w_{t,1},\dots,w_{t,k-1},w_*)  \cdot \nonumber \\ p_{\overleftarrow{\text {LM}}}(w_*,w_{t,k+1},\dots,w_{t,l_t})\Big].
\end{align}
For word insertion, the top-$K$ vocabulary $\mathcal{W}_{t,\text{insert}}$ is computed in a similar way (except that the position of $w_*$ is slightly different). Details are not repeated. In our experiments, $K$ is set to 50.

\textbf{Copy Mechanism.}
We observe that name entities and rare words are sometimes deleted or replaced during SA stochastic sampling. They are difficult to be recovered because they usually have a low language model-suggested probability.

Therefore, we propose a copy mechanism for SA sampling, inspired by that in Seq2Seq learning~\cite{gu2016incorporating}. Specifically, we allow the candidate sentence generator to copy the words from the original sentence $\mathrm x_0$ for word replacement and insertion. This is essentially enlarging the top-$K$ sampling vocabulary with the words in $\mathrm x_0$, given by
\small
\begin{align}
\widetilde{\mathcal{W}}_{t,\text{op}} = \mathcal{W}_{t,\text{op}} \cup \{ w_{0,1},\dots,  w_{0,l_0}\};\ \ 
\text{op}\! \in\! \{\text{replace,insert}\}
\end{align}
\normalsize
{Thus, $\widetilde{\mathcal{W}}_{t,\text{op}}$ is the actual vocabulary from which SA samples the word $w_*$ for replacement and insertion operation. }

While such vocabulary reduces the proposal space, it works well empirically because other low-ranked candidate words are either irrelevant or makes the sentence influent; they usually have low objective scores, and are likely to be rejected even if sampled.

\begin{algorithm}[t]
	\caption{UPSA}\label{alg:upsa}\footnotesize
	\begin{algorithmic}
		\State \!\!\!\!1:\ \textbf{Input}: Original sentence $\mathrm x_0$
		\State \!\!\!\!2:\ \textbf{for} $t\in \{1,\dots,N\}$ \textbf{do}
		\State \!\!\!\!3:\ \ \ \ $T =\max\{T_{\text{init}}- C\cdot t,0\}$
		\State \!\!\!\!4:\ \ \ \ Randomly choose an editing operation and a position $k$
		\State \!\!\!\!5:\ \ \ \ Obtain a candidate $\mathrm x_*$ by candidate sentence generator
	\State \!\!\!\!6:\ \ \ \ Compute the accepting probability $p_{\text{accept}}$ by Eqn.~(\ref{eq:temp})
		\State \!\!\!\!8:\ \ \ \ With probability $p_{\text{accept}}$,  $\mathrm  x_{t+1} = \mathrm x_*$
		\State \!\!\!\!9:\ \ \ \ With probability $1-p_{\text{accept}}$,  $\mathrm x_{t+1} = \mathrm x_t$
		\State \!\!\!\!\!\!\!10:\ \textbf{end for}
		\State \!\!\!\!\!\!\!10:\ \textbf{return} $\mathrm x_\tau$ s.t. $\tau=\text{argmax}_{\tau\in \{1,\dots,N\}} f(\mathrm x_\tau)$
\end{algorithmic}
\end{algorithm}

\subsection{Overall Optimization Process}
We summarize our simulated annealing algorithm for unsupervised paraphrasing (UPSA), also shown in Algorithm~\ref{alg:upsa}.

Given an input $\mathrm x_0$, UPSA searches from the sentence space to maximize our objective $f(\mathrm x)$, which involves semantic preservation, expression diversity, and language fluency. UPSA starts from $\mathrm x_0$ itself. For each searching step, it randomly selects a searching action (namely, word insertion, deletion, and replacement) at a position $k$ (Line~4); if insertion or replacement is selected, UPSA also proposes a candidate word, so that a candidate paraphrase $\mathrm  x_*$ is formed~(Line~5). Then, UPSA computes an acceptance rate $p_\text{accept}$ based on the increment of $f$ and the temperature $T$ (Line~6). The candidate sentence $\mathrm  x_{t+1}$ for the next step becomes $\mathrm  x_t$ if the proposal is accepted, or remains $\mathrm x_t$ if the proposal is rejected. Until the maximum searching iterations, we choose the sentence $\mathrm x_\tau$ that yields the highest score.

\section{Experiments}

\label{s:dataset}
\subsection{Datasets}  

   \textbf{Quora.} The Quora question pair dataset (Footnote 2) contains 140K parallel sentences and additional 640K non-parallel sentences. We follow the unsupervised setting in~\citet{miao2018cgmh}, where there are 3K and 20K pairs for validation and test, respectively. 
   
   	 \textbf{Wikianswers.} The original Wikianswers dataset~\cite{fader2013paraphrase} contains 2.3M pairs of question paraphrases from the Wikianswers website.\footnote{\url{http://knowitall.cs.washington.edu/paralex/}}
   	 Since our model only involves training a language model, we randomly selected 500K non-parallel sentences for training. For evaluation, we followed the same protocol as \citet{zichao2019} and randomly sampled 5K for validation and 20K for testing. Although the exact data split in previous work is not available, our results are comparable to previous ones in the statistical sense.
   	 
  \textbf{MSCOCO.} The MSCOCO dataset contains 500K+ paraphrases pairs for $\sim$120K image captions~\cite{lin2014microsoft}. We follow the standard split~\cite{lin2014microsoft} and the evaluation protocol in \citet{prakash2016neural} where only image captions with fewer than 15 words are considered. 
  
  \textbf{Twitter.} The Twitter URL paraphrasing corpus~\cite{lan2017a} is originally constructed for paraphrase identification. We follow the standard train/test split, but take 10\% of the training data as the validation set. The remaining samples are used to train our language model. For the test set, we only consider sentence pairs that are labeled as ``paraphrases.'' This results in 566 test cases.

\subsection{Competing Methods and Metrics} 

Unsupervised paraphrasing is an emerging research topic, and we could only find two plausible competing methods (namely, VAE and CGMH) in this setting. Early work on unsupervised paraphrasing typically adopts rule-based methods~\cite{mckeown1983paraphrasing,Barzilay-rules}. Their performance could not be verified on the above datasets, since the extracted rules are not available. Therefore, we are unable to compare them in this paper. Also, rule-based systems usually do not generalize well to different domains. In the following, we describe our competing methods:

$\bullet$ \textbf{VAE.} We train a variational autoencoder (VAE) with two-layer, 300-dimensional LSTM units.\footnote{We used the code in https://github.com/timbmg/Sentence-VAE} The VAE is trained with non-parallel corpora by maximizing the variational lower bound of log-likelihood; during inference, sentences are sampled from the learned variational latent space~\cite{bowman2015generating}.

$\bullet$ \textbf{CGMH.} \citet{miao2018cgmh} use Metropolis--Hastings sampling in the word space for constrained sentence generation. It is shown to outperform latent space sampling as in VAE, and is the state-of-the-art unsupervised paraphrasing approach. We adopted the published source code and generated paraphrases for comparison. 

We further compare UPSA with supervised Seq2Seq paraphrase generators: ResidualLSTM~\cite{prakash2016neural}, VAE-SVG-eq~\cite{gupta2018deep}, Pointer-generator~\cite{see2017get}, the Transformer~\cite{vaswani2017attention}, and the decomposable neural paraphrase generation (DNPG, \citeauthor{zichao2019} \citeyear{zichao2019}). DNPG has been reported as the state-of-the-art supervised
paraphrase generator.

To better compare UPSA with all paraphrasing settings, we also include domain-adapted supervised paraphrase generators that are trained in a source domain but tested in a target domain, including shallow fusion~\cite{gulcehre2015using} and
multi-task learning (MTL, \citeauthor{domhan2017using} \citeyear{domhan2017using}).

We adopt BLEU~\cite{papineni2002bleu} and ROUGE~\cite{lin2004rouge} scores as automatic metrics to evaluate model performance. \citet{sun2012joint} observe that BLEU and ROUGE could not measure the diversity between the generated and the original sentences, and propose the iBLEU variant by penalizing by the similarity with the original sentence.
Therefore, we regard the iBLEU score as our major metric, which is also adpoted in \citet{zichao2019}.

In addition, we also conduct human evaluation in our experiments (detailed later).

\subsection{Implementation Details}  
Our method involves unsupervised language modeling (forward and backward), realized by two-layer LSTM with 300 hidden units and trained specifically on each dataset with non-parallel sentences.

For hyperparameter tuning, we applied a grid search procedure on the validation set of the Quora dataset using the iBLEU metric. The power weights $P, Q,$ and $S$ in the objective were 8, 1, and 1, respectively,  chosen from $\{0.5,1,2,\dots,8\}$.

The initial temperature $T_\text{init}$ was chosen from $\{0.5,1,3,5,7,9\}\times10^{-2}$ and set to $T_{\text{init}}=3\times10^{-2}$ by validation. The magnitude of $T_\text{init}$ appears small here, but is in fact dependent on the scale of the objective function. 
The annealing rate $C$ was set to $\frac{T_{\text{init}}}{\#\text{Iteration}}=3\times10^{-4}$, where our number of iterations (\#\text{Iteration}) was $100$.

We should emphasize that all SA hyperparameters were validated only on the Quora dataset, and we did not perform any tuning on the other datasets (except the language model). This shows the robustness of our UPSA model and its hyperparameters.

\begin{table*}[t]	
	\begin{center}
	\small
	\resizebox{0.85\linewidth}{!}{
		\begin{tabular}{llcccccccc}
			\hline\noalign{\smallskip}
				   \multirow{2}*{} & & \multicolumn{4}{c}{Quora} &\multicolumn{4}{c}{Wikianswers}\\
		    \cmidrule(r){3-6}\cmidrule(r){7-10}
			
		    & Model & iBLEU & BLEU  & Rouge1 &  Rouge2 & iBLEU & BLEU  & Rouge1 &  Rouge2 \\
			\hline
			\noalign{\smallskip}
			\multirow{6}*{Supervised} 
			& ResidualLSTM      & 12.67 & 17.57 & 59.22 & 32.40 & 22.94 & 27.36  & 48.52 & 18.71\\
			& VAE-SVG-eq        & 15.17 & 20.04 & 59.98 & 33.30 & 26.35  & 32.98 & 50.93 & 19.11\\
			& Pointer-generator & 16.79 & 22.65 & 61.96 & 36.07 & 31.98 & 39.36  & 57.19 & 25.38   \\
    		& Transformer       & 16.25 & 21.73 & 60.25 & 33.45 & 27.70  & 33.01 & 51.85 & 20.70   \\
    		& Transformer+Copy  & 17.98 & 24.77 & 63.34 & 37.31  & 31.43 & 37.88 & 55.88 & 23.37 \\
    		& DNPG               & \underline{\textbf{18.01}} & \underline{\textbf{25.03}}  & \underline{\textbf{63.73}} & \underline{\textbf{37.75}} & \underline{\textbf{34.15}} & \underline{\textbf{41.64}}  & \underline{\textbf{57.32}} & \underline{\textbf{25.88}}  \\
			\noalign{\smallskip}
			\hline
			\noalign{\smallskip}
			\multirow{4}*{Supervised } 
			& Pointer-generator  & 5.04 &  6.96  & 41.89  & 12.77 & 21.87  & 27.94 & 53.99 & 20.85   \\
			& Transformer+Copy   & 6.17 &  8.15  & 44.89  & 14.79  & 23.25 & 29.22  & 53.33 & 21.02\\
			& Shallow fusion     & 6.04 &  7.95 & 44.87  & 14.79   &22.57  & 29.76 & 53.54 &
 20.68 \\
			\multirow{1}*{+ Domain-adapted} 			
			& MTL  & 4.90 &  6.37  & 37.64 & 11.83 & 18.34 & 23.65  & 48.19 & 17.53 \\
			& MTL+Copy  & 7.22 & 9.83  & 47.08 & 19.03 &21.87 & 30.78 & 54.10 & 21.08 \\
			& DNPG  & \underline{10.39} &  \underline{16.98}  & \underline{56.01} & \underline{28.61}  & \underline{25.60} & \underline{35.12} & \underline{56.17} & \underline{23.65} \\
			\noalign{\smallskip}
			\hline
			\noalign{\smallskip}
			\multirow{3}*{Unsupervised } 
			& VAE   & 8.16 &  13.96  & 44.55 & 22.64 & 17.92 & 24.13  & 31.87 & 12.08 \\
			& CGMH  & 9.94 & 15.73   & 48.73 & 26.12 & 20.05 & 26.45  & 43.31  & 16.53 \\
			& UPSA  & \underline{12.02} & \underline{18.18}   & \underline{56.51} & \underline{30.69} & \underline{24.84} & \underline{32.39}   & \underline{54.12} & \underline{21.45}  \\
			\noalign{\smallskip}
			\hline
		\end{tabular}
	}
	\end{center}\vspace{-1mm}
		\caption{Performance on the Quora and Wikianswers datasets. The results of supervised learning and domain-adapted supervised methods are quoted from~\citet{zichao2019}. We run experiments for all unsupervised methods and use the same evaluation script with~\citet{zichao2019} for a fair comparison. The results of CGMH in this table is slightly different from~\citet{miao2018cgmh}, because~\citet{miao2018cgmh} use corpus-level BLEU, while \citet{zichao2019} and we use sentence-level BLEU.}
	\label{table:wiki}
\end{table*}

\begin{table*}[t]
 \begin{center}
	\small
	\resizebox{0.6\linewidth}{!}{
		\begin{tabular}{ccccccccc}
		\hline
		    \noalign{\smallskip}
		   \multirow{2}*{Model} & \multicolumn{4}{c}{MSCOCO} &\multicolumn{4}{c}{Twitter}\\
		    \cmidrule(r){2-5}\cmidrule(r){6-9}
		& iBLEU	&  BLEU & Rouge1 & Rouge2
		& iBLEU	& BLEU  & Rouge1 & Rouge2\\
			\noalign{\smallskip}
		    \hline
			\noalign{\smallskip}
			\!\!\!VAE & 7.48 & 11.09    & 31.78 & 8.66     & 2.92 & 3.46  & 15.13 & 3.40   \\
			\!\!\!CGMH\!\!\!\!\!\! & 7.84  & 11.45  & 32.19    & 8.67    & 4.18   & 5.32   & 19.96 & 5.44   \\
			UPSA & \textbf{9.26} &  \textbf{14.16}    & \textbf{37.18} &  \textbf{11.21}  & \textbf{4.93} & \textbf{6.87}   &   \textbf{28.34} & \textbf{8.53}\\
			\hline
		\end{tabular}
		}
		\end{center}\vspace{-2mm}
	\caption{Performances on MSCOCO and Twitter.}\vspace{-2mm}
	\label{table:coco}
\end{table*}

\subsection{Results}

Table~\ref{table:wiki} presents the performance of all competing methods on the Quora and Wikianswers datasets. The unsupervised methods are only trained on the non-parallel sentences. The supervised models were trained on 100K paraphrase pairs for Quora and 500K pairs for Wikianswers. The domain-adapted supervised methods are trained on one dataset (Quora or Wikianswers) and tested on the other (Wikianswers or Quora).

We observe in Table~\ref{table:wiki} that, among unsupervised approaches,  VAE achieves the worst performance on both datasets, indicating that paraphrasing by latent space sampling is worse than word editing.
We further observe that UPSA yields significantly better results than CGMH: the iBLEU score of UPSA is higher than that of CGMH by 2--5 points. This shows that paraphrase generation is better modeled as an optimization process, instead of sampling from a distribution.

{It is curious to see how our unsupervised paraphrase generator is compared with supervised ones, should large-scale parallel data be available.} Admittedly, we see that supervised approaches generally outperform UPSA, as they can learn from massive parallel data. Our UPSA nevertheless achieves comparable results with the recent ResidualLSTM model~\cite{prakash2016neural}, reducing the gap between supervised and unsupervised paraphrasing. 

In addition, our UPSA could be easily applied to new datasets and new domains, whereas the supervised setting does not generalize well. This is shown by a domain adaptation experiment, where a supervised model is trained on one domain but tested on the other. 
We notice in Table~\ref{table:wiki} that the performance of supervised models (e.g., Pointer-generator and Transformer+Copy) decreases drastically on out-of-domain sentences, even if both Quora and Wikianswers are  question sentences. The performance is supposed to  decrease further if the source and target domains are more different.  UPSA outperforms all supervised domain-adapted paraphrase generators (except DNPG on the Wikianswers dataset), showing the generalizability of our model.

Table~\ref{table:coco} shows model performance on MSCOCO and Twitter corpora. These datasets are less widely used for paraphrase generation than Quora and Wikianswers, and thus, we only compare unsupervised approaches by running existing code bases. Again, we see the same trend as Table~\ref{table:wiki}: UPSA achieves the best performance, CGMH second, and the VAE worst.
It is also noted that the Twitter corpus yields lower iBLEU scores for all models, which is largely due to the noise of Twitter utterances~\cite{lan2017a}. However, the consistent results demonstrate that UPSA is robust and generalizable to different domains (without hyperparameter re-tuning).

\begin{table}[!t]
	\begin{center}
	\resizebox{\linewidth}{!}{
		\begin{tabular}{ccccccc}
			\hline\noalign{\smallskip}
		 \multirow{2}*{Model} & \multicolumn{2}{c}{Relevance} &\multicolumn{2}{c}{Fluency}\\
		    \cmidrule(r){2-3}\cmidrule(r){4-5}
			& Mean Score & Agreement  & Mean Score & Agreement \\
			\noalign{\smallskip}
			\hline
			\noalign{\smallskip}
			VAE   & 2.65 & 0.41 & 3.23  & 0.51 \\
		    CGMH  & 3.08 & 0.36  & 3.51 & 0.49 \\
			UPSA  & \textbf{3.78} & 0.55 & \textbf{3.66} & 0.53 \\
			\hline
		\end{tabular}
	}
	\end{center}\vspace{-2mm}
	\caption{Human evaluation on the Quora dataset.}\vspace{-3mm}
	\label{table:human}
\end{table}

\textbf{Human Evaluation.} We also conducted human evaluation on the generated paraphrases. Due to the limit of budget and resources, we sampled 300 sentences from the Quora test set and only compared the unsupervised methods (which is the main focus of our work). Selecting a subset of models and data samples is a common practice for human evaluation in previous work~\cite{wang2019a,li2017paraphrase}.

We asked three human annotators to evaluate the generated paraphrases in terms of relevance and fluency; each aspect was scored from $1$ to $5$. We report in Table~\ref{table:human} the average human scores and the Cohen's kappa score~\cite{cohen1960coefficient}.\footnote{According to~\citet{mchugh2012interrater}, a kappa score larger than 0.4 indicates moderate inter-annotator agreement.}
It should be emphasized that our human evaluation was conducted in a blind fashion.

Table~\ref{table:human} shows that UPSA achieves the highest human satisfaction scores in terms of both relevance and fluency. The results are also consistent with the automatic metrics in Tables~\ref{table:wiki} and~\ref{table:coco}.

We further conducted two-sided Wilcoxon signed rank tests. The improvement of UPSA is statistically significant with $p<0.01$ in both aspects, compared with both competing methods (UPSA vs.~CGMH and UPSA vs.~VAE).
\begin{table*}[t]\centering\footnotesize
	\resizebox{\linewidth}{!}{
		\begin{tabular}{|p{0.22\linewidth}|p{0.23\linewidth}|p{0.23\linewidth}|p{0.25\linewidth}|}
		\hline
		\multicolumn{1}{|c}{Input} & \multicolumn{1}{|c}{VAE} & \multicolumn{1}{|c}{CGMH}    & \multicolumn{1}{|c|}{UPSA}  \\
		\hline
		what would you do if given the power to become invisible ? & what would you do given the power to be invisible ? (4.33) & what do you do if given more power ? (3.33) & what would you do when you have a power to be invisible ? (4.67) \\\hline
		how can i become good in studies ? & how can i have a good android phone ? (2.33) & how can i become very rich in studies ? (4.00) & how should i do to get better grades in my studies ? (4.33) \\\hline
	    what are the best colleges for mass communication in india ? & what are the best way of communication in india ? (2.67) & which are the top universities for mass marketing in india ? (3.67) & which is the top university for mass communication in india ? (4.33)\\\hline
		how does one avoid existential depression ? & how does one avoid belly fats ? (2.67) & how do i overcome my ocd ? (2.67) & how do you get over existential depression ? (4.33)\\\hline
		what are the pluses and minuses about life as a foreigner in singapore ? & what are the UNK and  most interesting life as a foreigner in medieval greece ? (2.33) & what are the misconception about UNK with life as a foreigner in western ? (2.33) & what are the mistakes and pluses life as a foreigner in singapore ?  (2.67) \\\hline
		\end{tabular}
		}
		\caption{Example paraphrases generated by different methods on the Quora dataset. The averaged score evaluated by three annotators is shown at the end of each generated sentence. }\vspace{-1mm}
	\label{table:example}
\end{table*}

\subsection{Model Analysis}
We analyze UPSA in more detail on the most widely-used Quora dataset, with a test subset of 2000 samples.

\textbf{Ablation Study.}  We first evaluate the searching objective function~(\ref{eq:obj}) in Lines 1--4 of Table~\ref{table:ablation}. The results show that each component of our objective (namely, keyword similarity, sentence similarity, and expression diversity) plays its role in  paraphrase generation.  

Line~5 of Table~\ref{table:ablation} shows the effect of our copy mechanism, which is used in word replacement and insertion. It yields roughly one iBLEU score improvement if we keep sampling those words in the original sentence.

Finally, we test the effect of the temperature decay in SA. Line~6 shows the performance if we fix the initial temperature during the whole searching process, which is similar to Metropolis--Hastings sampling.\footnote{The Metropolis--Hastings sampler computes its acceptance rate in a different way from Eqn.~(\ref{eq:temp}).} The result shows the importance of the annealing schedule. It also verifies our intuition that sentence generation (in particular, paraphrasing in this paper) should be better modeled as a searching problem than a sampling problem.

\begin{table}[!t]
	\begin{center}
	\resizebox{0.96\linewidth}{!}{
		\begin{tabular}{clccccc}
			\hline\noalign{\smallskip}
			Line \# & UPSA Variant & iBLEU  & BLEU & Rouge1  & Rouge2 \\
			\noalign{\smallskip}
			\hline
			\noalign{\smallskip}	
			1 & UPSA   & \textbf{12.41} & 18.48 & 57.06 & 31.39 \\
			\noalign{\smallskip}
			\hline
			\noalign{\smallskip}	
			2& w/o $f_{\text{sim,key}}$ & 10.28  & 15.34  & 50.85 & 26.42\\
			3 & w/o $f_{\text{sim,sen}}$ & 11.78 & 17.95  & 57.04 & 30.80\\
			4  & w/o $f_{\text{exp}}$   & 11.93 & 21.17 & 59.75 & 34.91 \\
			5 & w/o copy  & 11.42 & 17.25 & 56.09 & 29.73\\
			6 & w/o annealing & 10.56 & 16.52  & 56.02 & 29.25 \\
			\hline
		\end{tabular}
		}
	\end{center}\vspace{-1mm}
	\caption{Ablation study. }\vspace{-2mm}
	\label{table:ablation}
\end{table}

\begin{figure}[t]
\centering
\includegraphics[width=0.75\linewidth]{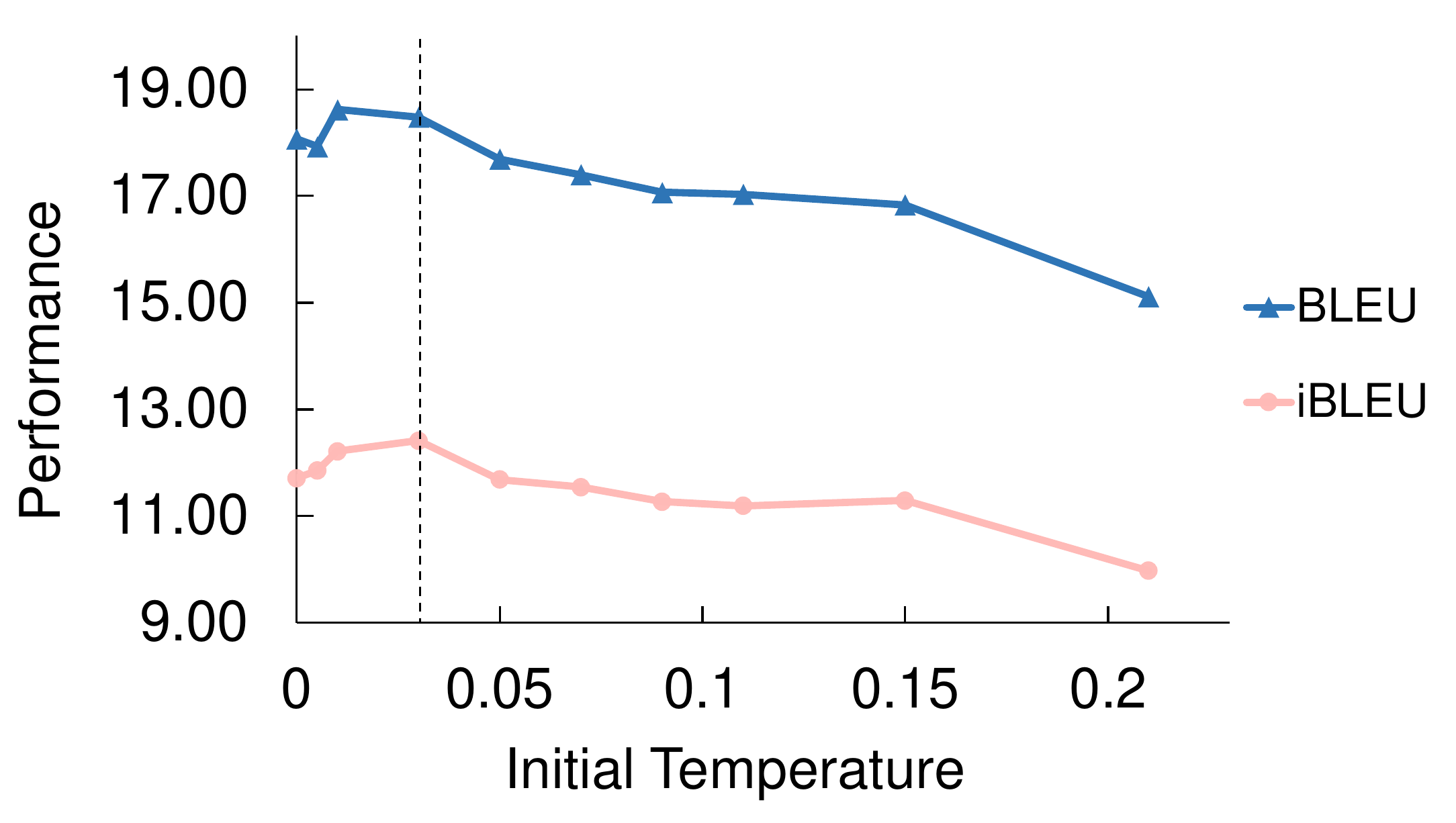}\vspace{-2mm}
\caption{Analysis of the initial temperature $T_{\text{init}}$. The dashed line illustrates the selected hyperparameter in validation.}
\label{fig:temp}
\end{figure}

\textbf{Analysis of the Initial Temperature.}  We fixed the decreasing rate to $C=1\times 10^{-4}$ and chose the initial temperature $T_\text{init}$ from $\{0,0.5,1,3,5,7,9,11,15,21\} \times 10^{-2}$. In particular, $T_\text{init}=0$ is equivalent to hill climbing (greedy search). The trend is plotted in Figure~\ref{fig:temp}.

It is seen that a high temperature yields worse performance (with other hyperparameters fixed), because in this case UPSA accepts more worse sentences and is less likely to settle down.

On the other hand, a low temperature makes UPSA greedier, also resulting in worse performance. Especially, our simulated annealing largely outperforms greedy search, whose temperature is 0.

We further observe that BLEU and iBLEU peak at different values of the initial temperature. 
This is because a lower temperature indicates a greedier strategy with less editing, and if the input sentence is not changed much, we may indeed have a higher BLEU score. Our major metric iBLEU penalizes the similarity to the input and thus prefers a higher temperature. We chose $T_\text{init}=0.03$ by validating on iBLEU. 

\textbf{Case Study.}
We showcase several generated paraphrases in Table~\ref{table:example}. We see qualitatively that UPSA produces more reasonable paraphrases than VAE and CGMH in terms of both closeness in meaning and difference in expressions, even for the relatively long sentences. For example, ``\textit{if given the power to become invisible}'' is paraphrased as ``\textit{when you have a power to be invisible}.''

From the examples, we also observe that our current UPSA mainly synthesizes a paraphrase by editing  words in the sentence, whereas the syntax of the original sentence is mostly preserved. This is partially due to the difficulty of exploring the entire (discrete) sentence space even by simulated annealing, and partially due to the insensitivity of the similarity objective  given two very different sentences.

\section{Conclusion}
In this paper, we proposed a novel unsupervised approach UPSA that generates a paraphrase of a given sentence by simulated annealing. We propose a searching objective function, involving semantic preservation, expression diversity, and language fluency. We also propose a copy mechanism as our searching action. Experiments on four benchmark datasets show that our model outperforms previous state-of-the-art unsupervised methods to a large extent. We further outperform most domain-adaptive paraphrase generators, as well as a supervised model on the Wikianswers dataset.

In the future, we plan to apply the SA framework on syntactic parse trees in hopes of generating more syntactically different sentences (motivated by our case study).

\fontsize{9.2pt}{10.1pt} \selectfont
\bibliography{ref}

\begin{thebibliography}{}

\bibitem[\protect\citeauthoryear{Anderson \bgroup et al\mbox.\egroup
  }{2017}]{anderson2016guided}
Anderson, P.; Fernando, B.; Johnson, M.; and Gould, S.
\newblock 2017.
\newblock Guided open vocabulary image captioning with constrained beam search.
\newblock In {\em EMNLP},  936--945.

\bibitem[\protect\citeauthoryear{Barzilay and Lee}{2003}]{Barzilay-rules}
Barzilay, R., and Lee, L.
\newblock 2003.
\newblock Learning to paraphrase: An unsupervised approach using
  multiple-sequence alignment.
\newblock In {\em ACL},  16--23.

\bibitem[\protect\citeauthoryear{Bowman \bgroup et al\mbox.\egroup
  }{2016}]{bowman2015generating}
Bowman, S.~R.; Vilnis, L.; Vinyals, O.; Dai, A.~M.; Jozefowicz, R.; and Bengio,
  S.
\newblock 2016.
\newblock Generating sentences from a continuous space.
\newblock In {\em CoNLL},  10--21.

\bibitem[\protect\citeauthoryear{Cohen}{1960}]{cohen1960coefficient}
Cohen, J.
\newblock 1960.
\newblock A coefficient of agreement for nominal scales.
\newblock {\em Educational and Psychological Measurement} 20(1):37--46.

\bibitem[\protect\citeauthoryear{Domhan and Hieber}{2017}]{domhan2017using}
Domhan, T., and Hieber, F.
\newblock 2017.
\newblock Using target-side monolingual data for neural machine translation
  through multi-task learning.
\newblock In {\em EMNLP},  1500--1505.

\bibitem[\protect\citeauthoryear{Dong \bgroup et al\mbox.\egroup
  }{2019}]{dong2019editnts}
Dong, Y.; Li, Z.; Rezagholizadeh, M.; and Cheung, J. C.~K.
\newblock 2019.
\newblock {EditNTS}: An neural programmer-interpreter model for sentence
  simplification through explicit editing.
\newblock {\em ACL}  3393--3402.

\bibitem[\protect\citeauthoryear{Edelkamp and
  Schroedl}{2011}]{edelkamp2011heuristic}
Edelkamp, S., and Schroedl, S.
\newblock 2011.
\newblock {\em Heuristic Search: Theory and Applications}.
\newblock Elsevier.

\bibitem[\protect\citeauthoryear{Ellsworth and
  Janin}{2007}]{ellsworth2007mutaphrase}
Ellsworth, M., and Janin, A.
\newblock 2007.
\newblock Mutaphrase: Paraphrasing with framenet.
\newblock In {\em Proc. {ACL}-{PASCAL} Workshop on Textual Entailment and
  Paraphrasing},  143--150.

\bibitem[\protect\citeauthoryear{Fader, Zettlemoyer, and
  Etzioni}{2013}]{fader2013paraphrase}
Fader, A.; Zettlemoyer, L.; and Etzioni, O.
\newblock 2013.
\newblock Paraphrase-driven learning for open question answering.
\newblock In {\em ACL},  1608--1618.

\bibitem[\protect\citeauthoryear{Granville, Krivanek, and
  Rasson}{1994}]{granville1994simulated}
Granville, V.; Krivanek, M.; and Rasson, J.
\newblock 1994.
\newblock Simulated annealing: a proof of convergence.
\newblock {\em IEEE Transactions on Pattern Analysis and Machine Intelligence}
  16(6):652--656.

\bibitem[\protect\citeauthoryear{Gu \bgroup et al\mbox.\egroup
  }{2016}]{gu2016incorporating}
Gu, J.; Lu, Z.; Li, H.; and Li, V.~O.
\newblock 2016.
\newblock Incorporating copying mechanism in sequence-to-sequence learning.
\newblock In {\em ACL},  1631--1640.

\bibitem[\protect\citeauthoryear{Gulcehre \bgroup et al\mbox.\egroup
  }{2015}]{gulcehre2015using}
Gulcehre, C.; Firat, O.; Xu, K.; Cho, K.; Barrault, L.; Lin, H.-C.; Bougares,
  F.; Schwenk, H.; and Bengio, Y.
\newblock 2015.
\newblock On using monolingual corpora in neural machine translation.
\newblock {\em arXiv preprint arXiv:1503.03535}.

\bibitem[\protect\citeauthoryear{Gupta \bgroup et al\mbox.\egroup
  }{2018}]{gupta2018deep}
Gupta, A.; Agarwal, A.; Singh, P.; and Rai, P.
\newblock 2018.
\newblock A deep generative framework for paraphrase generation.
\newblock In {\em AAAI},  5149--5156.

\bibitem[\protect\citeauthoryear{Guu \bgroup et al\mbox.\egroup
  }{2018}]{guu2018generating}
Guu, K.; Hashimoto, T.~B.; Oren, Y.; and Liang, P.
\newblock 2018.
\newblock Generating sentences by editing prototypes.
\newblock {\em Transactions of the Association for Computational Linguistics}
  6:437--450.

\bibitem[\protect\citeauthoryear{Hwang}{1988}]{hwang1988simulated}
Hwang, C.-R.
\newblock 1988.
\newblock Simulated annealing: theory and applications.
\newblock {\em Acta Applicandae Mathematicae} 12(1):108--111.

\bibitem[\protect\citeauthoryear{Kirkpatrick, Gelatt, and
  Vecchi}{1983}]{Kirkpatrick671}
Kirkpatrick, S.; Gelatt, C.~D.; and Vecchi, M.~P.
\newblock 1983.
\newblock Optimization by simulated annealing.
\newblock {\em Science} 220(4598):671--680.

\bibitem[\protect\citeauthoryear{Knight and Marcu}{2000}]{knight2000statistics}
Knight, K., and Marcu, D.
\newblock 2000.
\newblock Statistics-based summarization step one: Sentence compression.
\newblock In {\em AAAI},  703--710.

\bibitem[\protect\citeauthoryear{Lan \bgroup et al\mbox.\egroup
  }{2017}]{lan2017a}
Lan, W.; Qiu, S.; He, H.; and Xu, W.
\newblock 2017.
\newblock A continuously growing dataset of sentential paraphrases.
\newblock In {\em EMNLP},  1224--1234.

\bibitem[\protect\citeauthoryear{Li \bgroup et al\mbox.\egroup
  }{2018}]{li2017paraphrase}
Li, Z.; Jiang, X.; Shang, L.; and Li, H.
\newblock 2018.
\newblock Paraphrase generation with deep reinforcement learning.
\newblock In {\em EMNLP},  3865--3878.

\bibitem[\protect\citeauthoryear{Li \bgroup et al\mbox.\egroup
  }{2019}]{zichao2019}
Li, Z.; Jiang, X.; Shang, L.; and Liu, Q.
\newblock 2019.
\newblock Decomposable neural paraphrase generation.
\newblock In {\em ACL},  3403--–3414.

\bibitem[\protect\citeauthoryear{Lin \bgroup et al\mbox.\egroup
  }{2014}]{lin2014microsoft}
Lin, T.; Maire, M.; Belongie, S.~J.; Hays, J.; Perona, P.; Ramanan, D.; Dollar,
  P.; and Zitnick, C.~L.
\newblock 2014.
\newblock Microsoft {COCO}: Common objects in context.
\newblock In {\em ECCV},  740--755.

\bibitem[\protect\citeauthoryear{Lin}{2004}]{lin2004rouge}
Lin, C.-Y.
\newblock 2004.
\newblock Rouge: A package for automatic evaluation of summaries.
\newblock In {\em Proc. Workshop on Text Summarization Branches Out},  74--81.

\bibitem[\protect\citeauthoryear{Lowerre and Reddy}{1980}]{beam}
Lowerre, B.~T., and Reddy, D.~R.
\newblock 1980.
\newblock The harpy speech recognition system.
\newblock In {\em Trends in Speech Recognition}.

\bibitem[\protect\citeauthoryear{McHugh}{2012}]{mchugh2012interrater}
McHugh, M.~L.
\newblock 2012.
\newblock Interrater reliability: The kappa statistic.
\newblock {\em Biochemia Medica} 22(3):276--282.

\bibitem[\protect\citeauthoryear{Mckeown}{1983}]{mckeown1983paraphrasing}
Mckeown, K.~R.
\newblock 1983.
\newblock Paraphrasing questions using given and new information.
\newblock {\em Computational Linguistics} 9(1):1--10.

\bibitem[\protect\citeauthoryear{Metropolis \bgroup et al\mbox.\egroup
  }{1953}]{metropolis1953equation}
Metropolis, N.; Rosenbluth, A.~W.; Rosenbluth, M.~N.; Teller, A.~H.; and
  Teller, E.
\newblock 1953.
\newblock Equation of state calculations by fast computing machines.
\newblock {\em The Journal of Chemical Physics} 21(6):1087--1092.

\bibitem[\protect\citeauthoryear{Miao \bgroup et al\mbox.\egroup
  }{2019}]{miao2018cgmh}
Miao, N.; Zhou, H.; Mou, L.; Yan, R.; and Li, L.
\newblock 2019.
\newblock Constrained sentence generation by {Metropolis--Hastings} sampling.
\newblock In {\em AAAI},  6834--6842.

\bibitem[\protect\citeauthoryear{Narayan, Reddy, and
  Cohen}{2016}]{narayan2016paraphrase}
Narayan, S.; Reddy, S.; and Cohen, S.~B.
\newblock 2016.
\newblock Paraphrase generation from latent-variable for semantic parsing.
\newblock In {\em INLG},  153--162.

\bibitem[\protect\citeauthoryear{Neidert \bgroup et al\mbox.\egroup
  }{2014}]{wmt14}
Neidert, J.; Schuster, S.; Green, S.; Heafield, K.; and Manning, C.
\newblock 2014.
\newblock {Stanford University'}s submissions to the {WMT} 2014 translation
  task.
\newblock In {\em Proc. 9th Workshop on Statistical Machine Translation},
  150--156.

\bibitem[\protect\citeauthoryear{Pagliardini, Gupta, and
  Jaggi}{2017}]{pagliardini2017unsupervised}
Pagliardini, M.; Gupta, P.; and Jaggi, M.
\newblock 2017.
\newblock Unsupervised learning of sentence embeddings using compositional
  n-gram features.
\newblock In {\em NAACL},  528--540.

\bibitem[\protect\citeauthoryear{Papineni \bgroup et al\mbox.\egroup
  }{2002}]{papineni2002bleu}
Papineni, K.; Roukos, S.; Ward, T.; and Zhu, W.-J.
\newblock 2002.
\newblock {BLEU}: a method for automatic evaluation of machine translation.
\newblock In {\em ACL},  311--318.

\bibitem[\protect\citeauthoryear{Pennington, Socher, and
  Manning}{2014}]{pennington2014glove}
Pennington, J.; Socher, R.; and Manning, C.
\newblock 2014.
\newblock {GloVe:} global vectors for word representation.
\newblock In {\em EMNLP},  1532--1543.

\bibitem[\protect\citeauthoryear{Prakash \bgroup et al\mbox.\egroup
  }{2016}]{prakash2016neural}
Prakash, A.; Hasan, S.~A.; Lee, K.; Datla, V.; Qadir, A.; Liu, J.; and Farri,
  O.
\newblock 2016.
\newblock Neural paraphrase generation with stacked residual {LSTM} networks.
\newblock In {\em COLING},  2923--–2934.

\bibitem[\protect\citeauthoryear{Rose \bgroup et al\mbox.\egroup
  }{2010}]{rose2010automatic}
Rose, S.; Engel, D.; Cramer, N.; and Cowley, W.
\newblock 2010.
\newblock Automatic keyword extraction from individual documents.
\newblock {\em Text Mining: Applications and Theory} 1:1--20.

\bibitem[\protect\citeauthoryear{Roy and
  Grangier}{2019}]{roy-grangier-2019-unsupervised}
Roy, A., and Grangier, D.
\newblock 2019.
\newblock Unsupervised paraphrasing without translation.
\newblock In {\em ACL},  6033--6039.

\bibitem[\protect\citeauthoryear{See, Liu, and Manning}{2017}]{see2017get}
See, A.; Liu, P.~J.; and Manning, C.~D.
\newblock 2017.
\newblock Get to the point: Summarization with pointer-generator networks.
\newblock {\em arXiv preprint arXiv:1704.04368}.

\bibitem[\protect\citeauthoryear{Shah \bgroup et al\mbox.\egroup
  }{2018}]{shah2018building}
Shah, P.; Hakkani-T{\"u}r, D.; T{\"u}r, G.; Rastogi, A.; Bapna, A.; Nayak, N.;
  and Heck, L.
\newblock 2018.
\newblock Building a conversational agent overnight with dialogue self-play.
\newblock {\em arXiv preprint arXiv:1801.04871}.

\bibitem[\protect\citeauthoryear{Sun and Zhou}{2012}]{sun2012joint}
Sun, H., and Zhou, M.
\newblock 2012.
\newblock Joint learning of a dual {SMT} system for paraphrase generation.
\newblock In {\em ACL},  38--42.

\bibitem[\protect\citeauthoryear{Van~den Oord, Vinyals, and
  others}{2017}]{van2017neural}
Van~den Oord, A.; Vinyals, O.; et~al.
\newblock 2017.
\newblock Neural discrete representation learning.
\newblock In {\em NIPS},  6306--6315.

\bibitem[\protect\citeauthoryear{Vaswani \bgroup et al\mbox.\egroup
  }{2017}]{vaswani2017attention}
Vaswani, A.; Shazeer, N.; Parmar, N.; Uszkoreit, J.; Jones, L.; Gomez, A.~N.;
  Kaiser, {\L}.; and Polosukhin, I.
\newblock 2017.
\newblock Attention is all you need.
\newblock In {\em NIPS},  5998--6008.

\bibitem[\protect\citeauthoryear{{Wang} \bgroup et al\mbox.\egroup
  }{2019}]{wang2019a}
{Wang}, S.; {Gupta}, R.; {Chang}, N.; and {Baldridge}, J.
\newblock 2019.
\newblock A task in a suit and a tie: Paraphrase generation with semantic
  augmentation.
\newblock In {\em AAAI},  7176--7183.

\bibitem[\protect\citeauthoryear{Wieting and Gimpel}{2017}]{wieting2017paranmt}
Wieting, J., and Gimpel, K.
\newblock 2017.
\newblock {ParaNMT-50M}: Pushing the limits of paraphrastic sentence embeddings
  with millions of machine translations.
\newblock In {\em ACL},  451--462.

\bibitem[\protect\citeauthoryear{Zhao \bgroup et al\mbox.\egroup
  }{2009}]{zhao2009application}
Zhao, S.; Lan, X.; Liu, T.; and Li, S.
\newblock 2009.
\newblock Application-driven statistical paraphrase generation.
\newblock In {\em ACL},  834--842.

\end{thebibliography}
\bibliographystyle{aaai}
\end{document}